\documentclass[10pt, a4paper]{article}
\usepackage{lrec2016}
\usepackage{multibib}
\newcites{languageresource}{Language Resources}
\usepackage{graphicx}
\usepackage{float}

\usepackage{epstopdf}
\usepackage[latin1]{inputenc}

\usepackage{xstring}

\usepackage{booktabs} 
\usepackage{xcolor}
\usepackage{url}
\usepackage{balance}

\usepackage{hyperref}

\newcommand{\hdb}{\mbox{HappyDB}}

\newcommand{\nummoments}{$100,000$\ }

\let\OldUrlFont\UrlFont
\renewcommand{\UrlFont}{\footnotesize\OldUrlFont}

\title{\hdb: A Corpus of 100,000 Crowdsourced Happy Moments
\thanks{Part of the work was done while Asai, Evensen, and Xu were at Recruit Institute of Technology.}
}

\name{Akari Asai$^{*}$, Sara Evensen$^{\dagger}$, Behzad Golshan$^{\ddagger}$,
      Alon Halevy$^{\ddagger}$, Vivian Li$^{\ddagger}$ \\
      {\bf\large Andrei Lopatenko$^{\ddagger}$, Daniela Stepanov$^{\ddagger}$,
      Yoshihiko Suhara$^{\ddagger}$, Wang-Chiew Tan$^{\ddagger}$, Yinzhan Xu$^{\dagger}$}}

\address{$^{*}$Univ. of Tokyo, $^{\dagger}$MIT, $^{\ddagger}$Recruit Institute of Technology \\
         akari-asai@g.ecc.u-tokyo.ac.jp, \{sevensen, xyzhan\}@mit.edu,\\
         \{behzad, alon, vivian, andrei, daniela, suharay, wangchiew\}@recruit.ai}

\abstract{The science of happiness is an area of positive psychology concerned with understanding what
behaviors make people happy in a sustainable fashion. Recently, there has been interest in developing
technologies that help incorporate the findings of the science of happiness into users' daily lives by
steering them towards behaviors that increase happiness. With the goal of building technology that can
understand how people express their happy moments in text, we crowd-sourced \hdb, a corpus of 100,000
happy moments that we make publicly available. This paper describes \hdb\ and its properties, and
outlines several important NLP problems that can be studied with the help of the corpus. We also apply
several state-of-the-art analysis techniques to analyze \hdb. Our results demonstrate the need for
deeper NLP techniques to be developed which makes \hdb\ an exciting resource for follow-on research.
\newline 
\Keywords{science of happiness, positive psychology, happyDB corpus, crowdsourcing}}

\begin{document}

\maketitleabstract

\section{Introduction}
\label{sec:intro}
The science of happiness is an area of positive psychology that studies the factors that sustain
people's happiness over time~\cite{MS2011,BF2009,SL2008}. One of the interesting findings of the
field~\cite{ED99} is that while 50\% of our happiness is genetically determined, and only 10\% of it is
determined by our life circumstances (e.g., finances, job, material belongings), 40\% of our happiness
is determined by behaviors that are under our control. Examples of such behaviors include investing in
long-term personal relationships, bonding with loved ones, doing meaningful work, and caring for one's
body and mind. Consequently, positive psychologists have focused on  devising methods to steer people
towards those behaviors. Fostering happiness has also received attention at the national policy level --
in a recent interview~\cite{murthy16}  the U.S.\ Surgeon General claimed that fostering happiness is an
important priority as one of the main ways to prevent disease and live a longer, healthier life. 

Naturally, there has been recent interest to develop technologies that help users incorporate the
findings of the science of happiness into their daily lives. Current applications that pursue this goal
generally fall into one of the two categories: (1) applications that suggest relevant content to the
users based on their answers to a predefined set of questions \cite{TrackYourHappiness,Happify,Happier}
or (2) applications in which users can log their emotions in a journaling-style environment but that
content is available mostly for their own reflection \cite{Bliss,Mojo,DayOne}. 

Our work has been to develop a journal-like application where users express their happy moments using
their own language, thereby allowing for more nuance in their description of what makes them happy.
The ultimate goal of our app is that it should understand from the text which activities make the user
happy and who else participated in those happy moments. The app can then provide a useful visualization
of the user's happy moments, offer meaningful follow-up questions, and over time learn to suggest other
activities that may benefit the user. 

As we started working on this application we quickly realized that understanding the different aspects
of happy moments is a challenging NLP problem that has received very little attention to date. In order
to advance the state of the art for this problem, we set out to crowd-source \hdb, a corpus of 100,000
moments that we released publicly at \url{http://rebrand.ly/happydb}.

This paper describes the \hdb\
corpus, and outlines a few NLP problems that can be studied with it. We  describe the application of a
few state-of-the-art  analysis techniques to the corpus resulting in several observations. We also 
discuss some additional annotations that we provide along with \hdb\ that would be useful to anyone
who wants to explore the corpus further. The upshot of all these
analyses, however, is that there is a  need for deeper NLP techniques in the analysis of happy moments
(and of emotions expressed in text in general), and thus \hdb\ provides an exciting opportunity for
follow-up research. 

In addition to the applications motivating our work, there are other areas in which a deeper
understanding of happy moments can be useful. Of particular note is the analysis (by advertisers or
third parties) of the sources of happiness relating to products and services from comments on social
media. Viewed in that perspective, analyzing happy moments can also be seen as a refined analysis of
sentiments (e.g., \cite{liu2012sentiment,INR-011}).

\hdb\ is a collection of sentences in which crowd-workers answered the question: {\em what made you
happy in the past 24 hours} (or alternatively, the past 3 months). Naturally, the descriptions of happy
moments exhibit a high degree of linguistic variation. Note that \hdb\ is {\em not} a longitudinal dataset
that follows individuals over a period of time. Some examples of happy moments are: 

\begin{enumerate}
    \item {\em My son gave me a big hug in the morning when I woke him up.}
    \vspace{-2mm}
    \item {\em I finally managed to make 40 pushups.}
    \vspace{-2mm}
    \item {\em I had dinner with my husband.}
    \vspace{-2mm}
    \item {\em Morning started with the chirping of birds and the pleasant sun rays.}
    \vspace{-2mm}
    \item {\em The event at work was fun. I loved spending time with my good friends and laughing.}
    \vspace{-2mm}
    \item {\em I went to the park with the kids. The weather was perfect!}
\end{enumerate}

Fully understanding a happy moment is obviously a problem that goes beyond natural language processing
into the fields of psychology and philosophy. Here we take an NLP perspective on the problem and set a
goal of understanding which activities happened in the happy moment and who participated in those activities.
Evaluating which of these activities is the true cause of happiness adds another level of complexity. 
For example, even for the very simple happy moment {\em I had dinner with my husband}, the extracted
activities could be ``having dinner'', ``being with the husband'', or something that is not explicitly
in the text such as ``having a date night without the children''. 

The following are several NLP-related problems that could be studied using \hdb. 

\begin{enumerate}
    \item What are the activities described in a given happy moment? What other components besides
    activities are important in the happy moment?  Which of these aspects are most central to the happy
    moment? 
    \item Can we discover common paraphrasings to describe activities that appear in happy moments?
    \item Can we discover whether the cause of happiness in a particular happy moment is related to the
    expectation the person had? For example, a happy moment can be written as {\em I got to spend time
    with my son} versus {\em I spent time with my son}. In the first case it seems that the person was
    partially happy because they didn't expect to be able to spend time with their son.   
    \item Can we reliably remove extraneous text in a happy moment? For example can we transform,
    ``{\em I am happy to hear that my friend is pregnant}'' to ``{\em My friend is pregnant}''. Note
    that removing extraneous information can be very helpful in understanding which activity or event is
    the cause of happiness.
    \item Can we create a useful ontology of activities that cause happiness and map happy moments onto
    that ontology. Such an ontology can be an important tool for recommending additional activities to
    the user. 
\end{enumerate}

Solutions to the questions raised above will require advances in NLP. In particular, we need techniques
that go beyond analysis of the happy moments at the keyword level and perform deeper analyses such as
semantic role labeling (into possibly a set of frames that leverage Framenet, Verbnet, and/or Propbank).
Further analysis also needs to accommodate ungrammatical sentences such as ``{\em Early morning in the
beach, having breakfast with the family.}''

In this paper, we lay the groundwork for a deeper exploration of \hdb. We begin by describing how \hdb\
was collected and cleaned. We present some basic statistics about \hdb\ demonstrating that it is a broad corpus.
We compare the topics and the emotional content of our corpus with other corpora using standard
state-of-the-art annotations which we are releasing with \hdb. We also illustrate another interesting
aspect of \hdb: moments describing experiences from the last 24 hours are significantly different from
those describing experiences from the last 3 months. Finally, we address the most basic research
problem at the heart of \hdb: classifying happy moments into categories. We show that even this problem
is extremely challenging as it is closely related to the problem of mining expressed emotions in short
sentences. We describe a set of crowd-sourced category annotations that will facilitate future research in the problem.

\section{\hdb: 100,000 happy moments :)}
\label{sec:happydb}

\begin{figure}
\noindent
\begin{center}
\includegraphics[height=150pt]{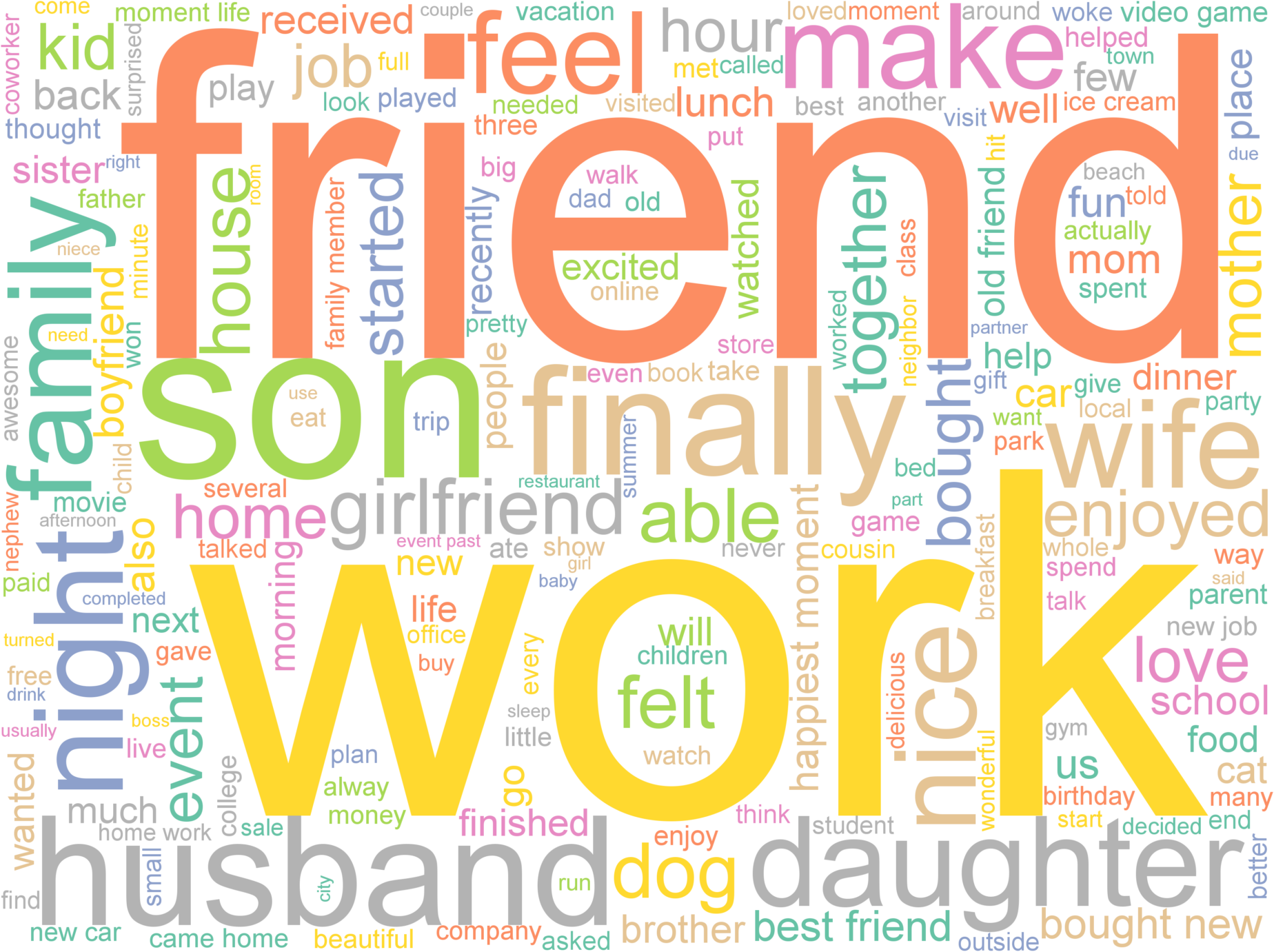}
\end{center}
\caption{\hdb's word cloud -- an anecdotal overview of the corpus. The words ``work'' and ``friend'' appear most
prominently in \hdb; mentions of ``wife'' and ``husband'' occur about equally, and so  do ``son''
and ``daughter''. However, ``girlfriend'' occurs more often than ``boyfriend'' (1960 vs. 1252 times),
``night'' appears more often than ``morning'' (3391 vs. 2736 times), and ``dog'' occur much more often
than ``cat'' (2160 vs. 988 times).}
\label{fig:wordcloud-distribution}
\end{figure}

We collected 100,000 happy moments with Mechanical Turk (MTurk) over 3 months. The workers were asked
to answer either: {\em what made you happy in the last 24 hours?} or, {\em what made you happy in the
last 3 months?} \hdb\ is split evenly between these two reflection periods. The majority of our workers
are of age 20 to 40 years and from the USA. There are about the same number of male and female workers
and the majority of our workers are single. More information about the demographics of the workers
as well as our crowd-sourcing setup can be found in appendices~\ref{sec:demographics} and~\ref{sec:datacollection}
respectively.
Along with the original $\nummoments$ happy moments (which
we refer to as the {\em original} \hdb), we also released a cleaned version of \hdb\ (which we refer to
as the {\em cleaned} \hdb), where some spelling mistakes are corrected (as described below) and some
vacuous moments are removed. Each moment is also annotated with the reflection period (24 hours or 3
months) and with the demographic information of the worker providing it. 

\begin{table}
\centering
\small
\begin{tabular}{|c|c|c|}\hline
{\bf Collection period} & 3/28/2017 -- 6/16/2017 \\\hline
{\bf \# happy moments} & 100,922 \\ \hline
{\bf \# distinct workers} & 10,843 \\ \hline
{\bf \# distinct words} & 38,188 \\ \hline
{\bf Avg. \# happy moments / worker} & 9.31 \\\hline
{\bf Avg. \# words / happy moment} & 19.66 \\\hline
\end{tabular}
\caption{Basic statistics on \hdb}\label{tab:dataset}
\normalsize
\end{table}

\smallskip
\noindent
{\bf Cleaning \hdb:~}
Naturally, the collected happy moments can contain a variety of errors. In our cleaning process we
dealt with two types of errors: (1) empty or single word sentences and (2) sentences with spelling
errors. We removed any sentences with less than two words. To find the spelling errors, we compared all
the words to a dictionary built from Norvig's text corpus~\cite{norvig2007write} as well as a complete
list of English Wikipedia titles\footnote{\url{https://dumps.wikimedia.org/enwiki/}} which includes the name of many cities, locations and
other known entities. We also performed a few edits on the dictionary to remove foreign language phrases
as well as certain words such as `Alot' and `Iam' which are actual city names,
but are more likely to be spelling errors. We found that only 2.7\% of happy moments contain
words not present in our dictionary. While this number seems small enough to justify removing such
happy moments, we observe that certain words are more likely to be misspelled and could create a bias
if we remove these happy moments. A specific example is that mentions of the word ``son" is higher than
``daughter'' in the original corpus because the word ``daughter'' is more likely to be misspelled than
the word ``son''. After fixing the typos using our technique (which we describe next), both words ended
up having almost the same frequency. This example indicates that there is a need for the spell-corrector.

To fix the spelling issues, we experimented with various open-source spell correctors, but we didn't 
find them suitable for our task; they either didn't provide confidence scores for the corrections,
or suggested corrections that would have a higher likelihood in other corpora, but not ours. For example,
in the context of happy moments, the phrase ``achive'' is more likely to be a typo for ``achieve'' than for
``active''. Thus, we decided to develop a spell corrector that is tailored to the domain of \hdb\ and only
corrects typos that we are highly confident of. The details of our spell-corrector are presented
in appendix~\ref{sec:spell}.

\begin{figure}
    \centering
    \includegraphics[trim={0 0 0 0},clip,width=230px]{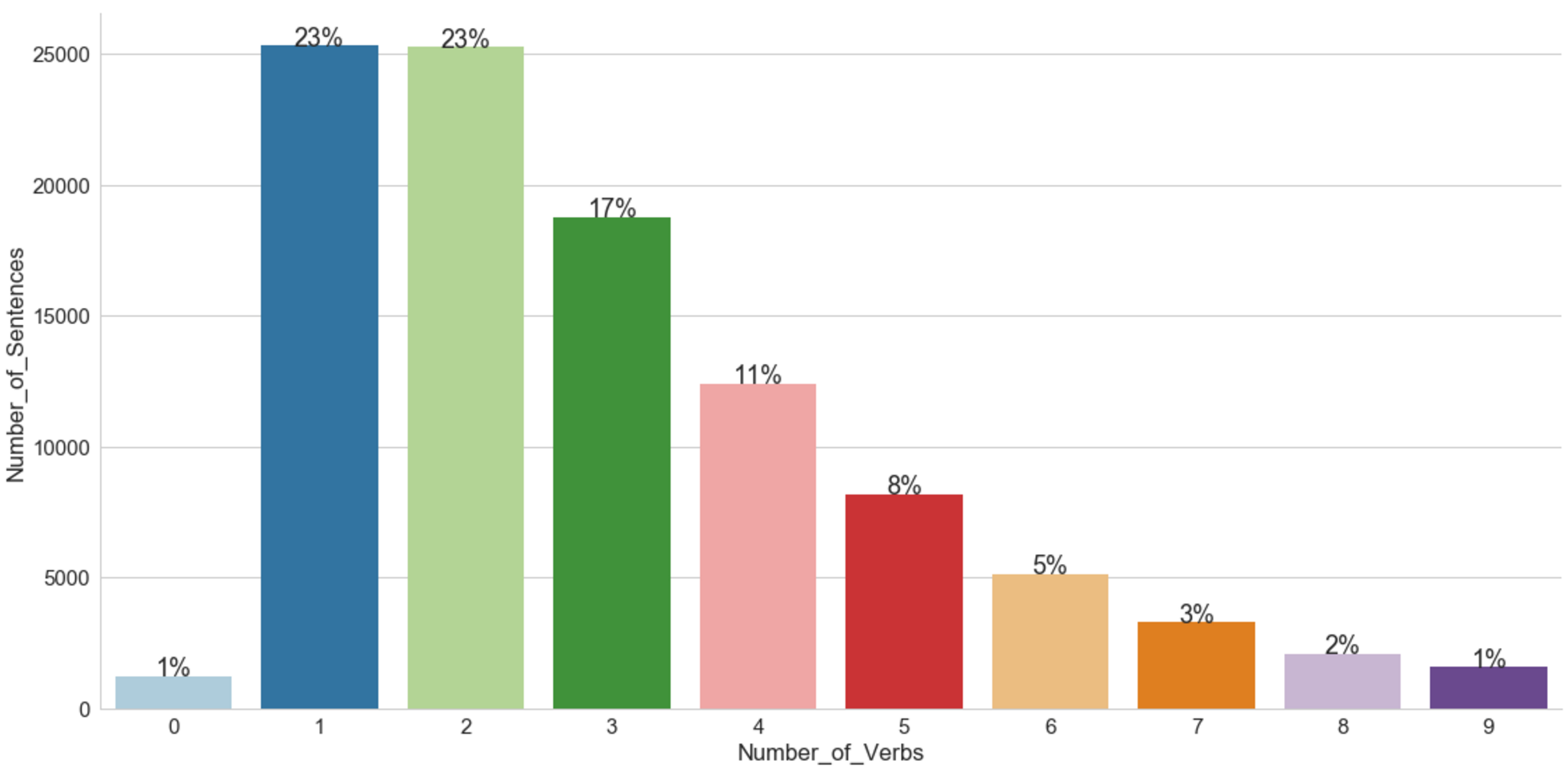}
    \caption{The distribution of the number of verbs per happy moment}
    \label{fig:verb_count}
\end{figure}

\smallskip
\noindent
{\bf Some basic statistics:~}
Table~\ref{tab:dataset} shows some basic statistics of the original \hdb.  
Figure~\ref{fig:wordcloud-distribution} shows the word cloud for the cleaned \hdb. The figure
is mostly provided for anecdotal value and as a means to highlight the most frequent words
in the corpus. As one proxy for the complexity of the sentences in \hdb, we calculated the number of verbs in each
sentence which are summarized in Figure~\ref{fig:verb_count}. The data shows that 53\% of the sentences
have 3 verbs or more and 36\% of the sentences have 4 verbs or more meaning that workers definitely
expressed quite complex thoughts in their moments.

\smallskip
\noindent
{\bf Diversity of contents in \hdb:~} An important question concerning the utility of \hdb\ is whether it
covers happy moments from a variety of topics. To get a feel for the level of diversity, we identified 9
rather diverse topics we saw occurring often in the corpus. The topics are ``people'', ``family'' (a
subset of ``people''), ``pets'', ``work'', ``food'', ``exercise'', ``shopping'', ``school'', and
``entertainment''. For each topic, we curated a list of keywords and regular expressions whose usage is
almost exclusive to the topic. For example, the category ``people'' contains words describing family members,
as well as other words that refer to people, like ``hairdresser'' or ``neighbor'', but it does not include
``he'', ``she'' or ``they'', as these words are sometimes used in reference to pets or inanimate objects.
Additionally, if these pronouns refer to a person they should also have an antecedent which our dictionary
should recognize. ``People'' also does not contain the word ``I'', since we are trying to capture interactions
between people.

Table~\ref{tab:topics} shows the percentage of sentences in \hdb\ found for each topic as well as the
size of the list associated with each topic. Note that a happy moment may be related to multiple topics.
For instance, ``\emph{running with my son}'' is related to both ``family'' and ``exercise''.
All of the keywords lists are disjoint except for
``people'': this is a superset of ``family'', and also contains some words from other topics,
for example ``co-worker'' which is also in ``work''. We can observe that 80\% of \hdb\ pertains to
these 9 topics. The remaining sentences that did not fit into any of these topics contain all sorts of
topics, such as rare surprises (``{\em finding a \$100 dollar bill inside my pants pockets}'') or
situations that turned out to be better than expected (``{\em There was almost no traffic today}'').
None of these other categories covered a large enough portion of our corpus to justify adding them
to our dictionaries. However, this suggests that there is a long tail of topics in the corpus.

\begin{table}
\centering
\small
\begin{tabular}{lrr}
\toprule
       &  \% of Sentences &  Size of \\
Topics &  in Topic        &  Keywords List\\
\midrule
people &         46.0   &  478 \\
family &         26.4   &  423 \\
food &           16.2   &  1073 \\
work &           14.5   &  115 \\
entertainment &  8.8    &  156\\
exercise &       8.4    &  558 \\
shopping &       8.4    &  35 \\
school &         5.5    &  47 \\
pets &           4.5    &  149 \\
none &           20.3   &  N/A \\
\bottomrule
\end{tabular}
\caption{\hdb\ Topics Distributions}
\label{tab:topics}
\normalsize
\end{table}

Another perspective on the contents of \hdb\ can be obtained by annotating the corpus with the popular
semantic classes known as {\em supersense}. Supersense tags are defined in WordNet
\cite{fellbaum1998wordnet} as lexicographic classes and are categorized into 15 verb classes (e.g.,
stative, cognition, communication, social, motion etc.) and 26 noun classes (e.g., person, artifact,
cognition, food etc.). 

\begin{table*}
\small
\centering
\begin{tabular}{l|lllllll}\hline
          & n.person        & n.time          & v.social        & v.motion        & v.possession    & n.event         & n.food          \\\hline\hline
HappyDB   & 0.48          & \textbf{0.41} & \textbf{0.31} & \textbf{0.26} & \textbf{0.22} & \textbf{0.18} & \textbf{0.13} \\
Blog      & 0.28          & 0.12          & 0.15          & 0.07           & 0.04          & 0.05          & 0.06          \\
Newspaper & 0.39          & 0.11          & 0.12          & 0.06          & 0.04          & 0.05          & 0.07          \\
Twitter   & \textbf{0.51} & 0.11          & 0.11          & 0.07          & 0.03          & 0.04          & 0.05         \\\hline
\end{tabular}
\caption{\small Distribution of top-7 most frequent supersense categories in HappyDB and other corpora. The number is
the proportion of sentences that contain a particular supersense label. Bold-face denotes the highest value in a
column.}
\label{table:supersense_dist}
\end{table*}

We trained a supersense tagger with the SemEval-2016 dataset \cite{Schneider:2015ul} using CRF
\cite{CRFsuite}. The supersense annotated \hdb\ is also provided as part of HappyDB. Table
\ref{table:supersense_dist} shows the proportion of sentences for the top seven supersense labels in
\hdb. It also displays the proportion of supersense labels for sentences in other textual corpora from
the Manually Annotated Sub-Corpus dataset\footnote{MASC v3.0.0 \url{http://www.anc.org/data/masc/downloads/data-download/}}.
As shown, the proportions of several of the top
five labels for \hdb\ are significantly higher than the other corpora which implies that these labels
are potential features for identifying happy moments. Examples of some supersense classes and their
frequencies in \hdb\ are shown in Table \ref{table:supersense_examples}.

\begin{table}
\centering
\small
\begin{tabular}{ll|ll|ll}\hline
n.person &       & n.food    &      & v.emotion &      \\\hline
friend   & 9,493 & food      & 1,511 & enjoy     & 2,778 \\
son      & 3,507  & dinner    & 938 & love      & 2,080 \\
daughter & 3,314  & coffee    & 681 & feel      & 1,399 \\
wife     & 2,721  & pizza      & 568  & like      & 1,070 \\
husband  & 2,687  & breakfast  & 525  & excite    & 533 \\\hline
\hline
\end{tabular}
\caption{Most common words associated with frequently used supersense labels.}
\label{table:supersense_examples}
\end{table}

\subsection{Emotions in happy moments}
To analyze the cognitive and emotional state of happy moments, we applied the sentiment lexicon
Linguistic Inquiry and Word Count (LIWC), ``a transparent text analysis program that counts words in
psychologically meaningful categories'', \cite{liwc} on a sample of 500 happy moments; only a sample
was chosen because of existing restrictions on the amount of requests to the LIWC commercial API.
Table~\ref{table:LIWC} shows some of the LIWC categories in which the
scores for the 500 happy moments vary notably from those of other corpora (expressive writing, blog
posts, and novels \cite{pennebaker2015development}). These categories are defined as follows:
\textit{analytic} refers to a measurement of the author's logical thinking, as opposed to narrative and
informal thinking; \textit{authentic} approximates how honest and disclosing the writing is; and
\textit{tone} measures how positive or negative the text is. As expected, \hdb\ has a higher score for
tone than any of the other corpora analyzed. More interestingly, the analytic score for \hdb\ is quite high
and very close to that of Novels, yet the authentic score (also quite high) is closer to that for Expressive
Writing. Our analysis of LIWC scores suggest that our corpus is very disclosing and honest which makes it an
ideal corpus for studying emotions expressed in text.

\begin{table}
\begin{center}
\small
\begin{tabular}{l|cccc}
\hline
{\bf LIWC} & & & & \\
{\bf Category} & {\bf \hdb} & {\bf Expressive} & {\bf Novels} & {\bf Blogs} \\
\hline
\hline
analytic &  69.19 & 44.88 & 70.33 & 49.89 \\
authentic & 77.14 & 76.01 & 21.56 & 60.93 \\
tone &      60.85 & 38.60 & 37.06 & 54.50 \\
\end{tabular}
\end{center}
\caption{Average LIWC Scores for (cleaned) \hdb\ compared to those of other text corpora of expressive
writing, novels, and blogs \protect\cite{pennebaker2015development}. All three categories provide scores in the
range from 1 to 99.}
\label{table:LIWC}
\end{table}

An alternative approach to analyze the emotions expressed in text is to use the
Valence-Arousal-Dominance model (VAD) of emotion~\cite{bradley1994measuring,warriner2013norms} which
provides a score for each lemmatized word on a scale of pleasure-displeasure (valence), excitement-calm
(arousal), and control-inhibition (dominance). To evaluate our data across these dimensions, we used the
Warriner et al. database of 13,915 manually rated English lemmas, as averaged over at least 18 ratings
for all three VAD features \cite{warriner2013norms}. This is currently the largest available lexicon of
VAD scores, and the VAD ratings covered 45.84\% of the lemmatized words in \hdb. The words which were
not covered were mostly pronouns, articles, conjunctions, numbers, and proper nouns.
Some examples of the highest and lowest scoring words
across each dimension are listed in Table \ref{table:warr}. We calculated a VAD score for \hdb\ by
taking the mean over the VAD score of words in the corpus. Interestingly, we observed that HappyDB's
VAD score is similar to the \emph{travel} section of the Guardian corpus~\cite{BP2013} (V$\cong$6.2,
A$\cong$4.0, D$\cong$5.7) and rather different from other sections such as \emph{crime} or
\emph{banking}. This shows that the VAD scores (which we release as part of HappyDB) can help us
quantify how emotional the content of the corpus is.

\begin{table}
\centering
\small
\begin{tabular}{l|l|l}
\toprule
{\bf Category} &   Low Scoring Words &   High Scoring Words \\
\midrule
Valence	&   murder (1.48)	&   excited (8.11)  \\ 
	    &   leukemia (1.47) &  	happiness (8.48)  \\
\hline
Arousal & librarian (1.75) &    rampage (7.57)  \\
	    & calm (1.67)      &    lover (7.45) \\
\hline
Dominance &	Alzheimer's (2.00) &    completion (7.73)  \\
          & earthquake (2.14)  &    smile (7.72) \\
\bottomrule
\end{tabular}
\caption{Examples of words with very high or low VAD scores~\protect\cite{warriner2013norms}.}
\label{table:warr}
\end{table}

The conclusion from the analyses provided so far in this section is that \hdb\ is a diverse corpus with
content that is emotionally rich and covers various topics (e.g., ``work'', ``leisure'', ``exercise'' and etc.).
Furthermore, while we used several techniques to extract general statistics about the content, diversity,
and emotional content of \hdb, there is clearly a need for deeper analysis of happy moments. 

\subsection{Comparing Reflection Periods}
The analyses presented thus far, though rather rudimentary, already enable us to discover an important
property of \hdb, namely that there are important differences between the happy moments that reflect on
the last 24 hours versus those that reflect 3 months back. In addition to being an important property
of the corpus, these  differences raise additional interesting research questions. We demonstrate these
differences in two ways. 

\smallskip
{\bf Pointwise Mutual Information Scores:~}
For each reflection period we calculated pointwise mutual information (PMI) scores \cite{IIR} for words
in the cleaned happy moments, and compared the top nouns in each batch. Table~\ref{tab:ratio} shows the top
10 nouns with the highest PMI scores in the 24 hours batch w.r.t.\ the other batch and vice-versa. The
results suggest that moments reported in the 24 hour period tend to be activities that occur daily
(e.g., foods, bedtime) and moments reported in the 3 months period tend to reflect infrequent
occurrences like holidays or life events.  

\begin{table}
\small
\begin{minipage}{0.25\textwidth}
\begin{tabular}{l|c}
\multicolumn{2}{c}{\bf 24 hours} \\
\hline
{\bf Word $w$} & {\bf Ratio} \\
& ($X_w/Y_w$)  \\
\hline
\hline
bedtime  &  15.0 \\
custard  &  12.0 \\
spoon    &     12.0 \\
burritos &    10.0 \\
nachos   &    10.0 \\
opener   &    10.0 \\
fool     &    10.0 \\
dough    &     9.0 \\
hurry    &     9.0 \\
gossip   &     9.0 \\
\hline
\end{tabular}    
\end{minipage}
\begin{minipage}{0.2\textwidth}
\begin{tabular}{l|c}
\multicolumn{2}{c}{\bf 3 months} \\
\hline
{\bf Word $w$} & {\bf Ratio} \\
& ($Y_w/X_w$) \\
\hline
\hline
valentine & 45.0 \\
scenario  & 34.0 \\
sorrow    & 24.0 \\
gender    & 20.0 \\
thousand  & 17.0 \\
custody   & 17.0 \\
faculty   & 16.0 \\
palace    & 14.0 \\
propose   & 13.0 \\
military  & 11.0 \\
\hline
\end{tabular}    
\end{minipage}
\caption{Top PMI words from the two batches. $X_w$ (resp. $Y_w$) denotes the probability of word
$w$ occurring in the 24 hours batch (resp. 3 months batch)}
\label{tab:ratio}
\end{table}

\smallskip
{\bf Topic Mentions by Reflection Period:~}
We analyzed the incidence of different topics separately for each reflection period. In Table
\ref{table:topic_reflection} we observe different distributions of topics for each reflection period,
mainly in the categories ``food'', ``school'', ``people'', ``family'', and ``entertainment''. For
instance, we observe that the categories ``food'' and ``entertainment'' have higher percentage of
coverage in 24 hour (19.2\%, and 9.6\%) compared to the 3 months reflection period (13.1\%, and 7.8\%).
Naturally, people are more likely to talk about a meal or a movie because these are more frequent daily
events that are more likely to be remembered if they occurred recently. When people are asked to
reflect on the past 3 months, they tend to remember events that are more prominent such as school, big
achievements, and time spent with friends and family.

\begin{table}
\small
\begin{center}
\begin{tabular}{l|r|r}
\hline
{\bf Topics} & \multicolumn{2}{c}{\bf \% of sentences in topic}  \\
 & {\bf 24 Hours Reflection} & {\bf 3 Months Reflection} \\
\hline
\hline
people &           44.0  &      47.9  \\
family &           24.7  &      27.9  \\
pets &             4.6   &      4.4   \\
work &             14.4  &      14.5  \\
food &             19.2  &      13.1  \\
exercise &         8.7   &      8.1   \\
shopping &         7.6   &      9.0   \\
school &           4.3   &      6.7   \\
entertainment &    9.6   &      7.8   \\
None &             20.4  &      20.1  \\
\hline
\end{tabular}
\end{center}
\caption{Percentage of happy moments in each topic, separated by reflection period. All differences
between the columns are statistically significant at $p < 10^{-3}$ except for pets, work, and none.}
\label{table:topic_reflection}
\end{table}

\label{sec:annotation}


\section{Categorizing Happy Moments}
\label{sec:categories}
So far we have gained an understanding of \hdb\ through some analysis of annotations that are frequently considered in the literature.
In this section, we take a first step towards a deeper analysis of happy moments by trying to classify them into categories.
Categorization is important for several reasons. First, it forms the basis for visualizing one's happy moments. Second, the techniques for analyzing happy
moments may depend partially on the category they belong to. Finally, the category of a happy moment could trigger a conversation between
an app and a user, and the course of conversation is clearly dependent on the category being discussed. For instance, the app's response 
to a happy moment about completing an exercise may be to congratulate the user, but the same response would be unacceptable if the user
mentions that she is enjoying a beautiful scenery. 

\begin{table*}
\small
\begin{tabular}{p{2.5cm}|p{7.5cm}|l}
\hline
{\bf Category} & {\bf Definition} & {\bf Examples} \\
\hline
\hline
Achievement & With extra effort to achieve a better than expected result &
Finish work. Complete marathon. \\
Affection & Meaningful interaction with family, loved ones and pets & Hug. Cuddle. Kiss.  \\
Bonding & Meaningful interaction with friends and colleagues & Have meals w coworker. Meet with friends. \\
Enjoy the moment & Being aware or reflecting on present environment & Have a good time. Mesmerize. \\
Exercise & With intent to exercise or workout & Run. Bike. Do yoga. Lift weights. \\
Leisure & An activity done regularly in one's free time for pleasure & Play games. Watch movie. Bake cookies. \\
Nature & In the open air, in nature & Garden. Beach. Sunset. Weather \\
\hline
\end{tabular}
\caption{The categories of happy moments} 
\label{tbl:categories}
\vspace*{-1em}
\end{table*}

There is no consensus on a single set of categories for happy moments in positive psychology because they are often
discussed under different names, with small variations and at different levels of granularity. We chose a set of
categories inspired by research in positive psychology that also reflects the contents of \hdb. These categories
and a brief description of them are listed in Table~\ref{tbl:categories}. Note that \emph{affection} refers to activity
with family members and loved ones, while \emph{bonding} refers to activities with other people in one's life.

We developed a multi-class classifier using Logistic Regression with a bag of words representation of happy 
moments as features. To obtain training data, we crowdsourced a batch of $15,000$ happy moments to obtain
category labels. 
Every happy moment was shown to 5 workers, and we only considered labels that at least 3 workers agreed on.
Table~\ref{tbl:classifyhappyDB} shows the performance of our classifier using a 5-fold cross-validation setup.
Clearly the classifier has room for further improvement on categories such as ``Leisure'' and ``Enjoy the moment''
which shows that word distributions are not sufficient for this task. Building a classifier with sufficiently high
precision/recall scores on all categories can be a challenging task in general, as it usually involves inferring 
some information or context that is not explicitly mentioned in text.

We publish our crowd-sourced labels
as part of \hdb\ to provide a ground-truth for researchers interested in topic mining and clustering of short utterances.
We also released our predicted results on the entire corpus as a baseline.

\begin{table}
\small
\begin{center}
\begin{tabular}{l@{\hspace{5pt}}|@{\hspace{2pt}}c@{\hspace{5pt}}c@{\hspace{5pt}}c@{\hspace{2pt}}|@{\hspace{2pt}}c@{\hspace{5pt}}c}
\hline
                &                 &              &          & \multicolumn{2}{c@{}}{\bf \% of moments} \\
 {\bf Category} & {\bf Precision} & {\bf Recall} & {\bf F1} & {\bf 24 Hrs} & {\bf 3 Months}\\
 \hline
 \hline
Achievement & 79.2 & 87.3 & 83.0 & 30.9 & 36.5\\
Affection & 89.9 & 94.3 & 92.0 & 32.7 & 35.1\\
Bonding & 91.9 & 87.1 & 89.4 & 10.4 & 10.8\\
Enjoy the moment & 59.2 & 49.9 & 54.0 & 13.3 & 8.9 \\
Exercise & 85.3 & 59.3 & 69.9 & 1.5 & 0.8\\
Leisure & 77.9 & 67.2 & 72.1 & 8.7 & 6.1\\
Nature & 80.9 & 52.4 & 63.4 & 2.1 & 1.5\\
\hline
\end{tabular}
\end{center}
\vspace*{-1em}
\caption{\small Precision/recall/F1 scores for each category as well as the \%
of moments per category for each reflection period. All differences between
the reflection periods are statistically significant at $p < 10^{-5}$ except for bonding.}
\vspace*{-1em}
\label{tbl:classifyhappyDB}
\end{table}

Table~\ref{tbl:classifyhappyDB} also shows the percentage of moments classified into each category for both
reflection periods, which  further highlights the differences between reflection periods described in Section~\ref{sec:happydb}.
Notice that \hdb\ has roughly the same number of moments for each reflection period. Thus, the percentage of moments
classified in each category in \hdb\ can be computed by taking the average of the last two columns in Table~\ref{tbl:classifyhappyDB}.
The higher frequency of moments in ``Exercise'', ``Nature'', and ``Leisure''
under the 24 hours reflection period confirms our theory that daily tasks are sources of short-term happiness.
Longer-term happiness is more likely to come from loved ones or achievements.

\section{Related work}
\label{sec:relatedwork}
To the best of our knowledge, \hdb\ is the first crowdsourced corpus of happy moments which can be used for understanding
the language people use to describe happy events. There has been recent interest in creating datasets in the area
of mental health.  Althoff et al.\ \cite{Althoff:2016ua} conducted a large scale analysis on counseling conversational 
logs collected from short message services (SMS) for mental illness study. They studied how
various linguistic aspects of conversations are correlated with conversation outcomes. Mihalcea et al.\ \cite{MihalceaL06}
performed text analysis on blog posts from LiveJournal (where posts can be assigned happy/sad tags by their authors).
Lin et al.\ \cite{HuijieLin:2016wy} measure stress from short texts by identifying the stressors and the
stress levels. They classified tweets into 12 stress categories defined by the stress scale in~\cite{Holmes:1967ea}.

Last year alone, there were multiple research efforts that obtained datasets via crowdsourcing and applied natural
language techniques to understand different corpora. For example, SQuAD~\cite{RZLL16} created a large-scale dataset 
for question-answering. The crowdsourced workers were asked to create
questions based on a paragraph obtained from Wikipedia. They employed
MTurk workers with strong experience and a high approval rating to ensure the quality of the dataset. 
We did not select the workers based on their qualification for \hdb\ as our task is cognitively easier than SQuAD's and 
we want to avoid bias in our corpus.
\hdb\ is similar to SQuAD in terms of the scale of the crowdsourced dataset. However, unlike SQuAD, which was designed
specifically for studying the question answering problem, the problems that \hdb\ can be used for are more open-ended.

\section{Conclusion}
\label{sec:conclusion}
We have published
\hdb, a broad corpus of happy moments expressed in diverse linguistic styles. We have also derived a cleaned version of \hdb,  added annotations, and presented our
analysis of \hdb\ based on these annotations. We made our dataset and most of our annotations publicly available to encourage further research in the science of happiness and well-being in general.
We believe that \hdb\ can spur research of the topic of understanding happy moments and more generally, the expression of emotions in text. The results of this research can 
translate to applications that can improve people's lives.

\balance
\section{Bibliographical References}
\bibliographystyle{lrec2016}
\bibliography{happypaper8pagenew}

\appendix

\section{Demographic information of \hdb\ crowdsourcing workers}
\label{sec:demographics}

The following tables represent the demographic distributions of the crowdsourcing workers who contributed to our \hdb\ dataset.

\begin{table}[h]
\centering
\begin{tabular}{lr}
\toprule
  Age &  Ratio \\
\midrule
 10--20 &  4.22\% \\
 20--30 & 47.77\% \\
 30--40 & 29.34\% \\
 40--50 & 10.22\% \\
 50--60 &  5.89\% \\
 60--70 &  2.15\% \\
 70--80 &  0.30\% \\
 80--90 &  0.04\% \\
\bottomrule
\end{tabular}
\caption{Age distribution.}
\end{table}

\begin{table}[h]
\centering
\begin{tabular}{lr}
\toprule
Country &  Ratio \\
\midrule
    USA & 86.11\% \\
    IND &  8.94\% \\
    CAN &  0.61\% \\
    VEN &  0.50\% \\
    GBR &  0.44\% \\
 OTHERS &  3.40\% \\
\bottomrule
\end{tabular}
\caption{Country of residence distribution.}
\end{table}

\begin{table}[h]
\centering
\begin{tabular}{lr}
\toprule
        Gender &  Ratio \\
\midrule
        Female & 50.37\% \\
          Male & 49.12\% \\
 Not specified &  0.51\% \\
\bottomrule
\end{tabular}
\caption{Gender distribution.}
\end{table}

\begin{table}[H]
\centering
\begin{tabular}{lr}
\toprule
     Marital status &  Ratio \\
\midrule
    Single & 52.66\% \\
   Married & 40.52\% \\
  Divorced &  5.22\% \\
 Separated &  0.95\% \\
   Widowed &  0.66\% \\
\bottomrule
\end{tabular}
\caption{Marital status distribution.}
\end{table}

\begin{table}[H]
\centering
\begin{tabular}{lr}
\toprule
Parenthood status &  Ratio \\
\midrule
               No & 59.64\% \\
              Yes & 40.36\% \\
\bottomrule
\end{tabular}
\caption{Parenthood status distribution.}
\end{table}

\section{Data collection by crowdsourcing}
\label{sec:datacollection}
We investigated several parameters before collecting our corpus. 
First, we investigated whether differences in our instructions to the workers will influence
the happy moments collected.
Second, we experimented with different windows of reflection (i.e., how far in the past did the happy moment occur). We did this to understand
how the time period influences the content of happy moments.
By analyzing the outcomes on batches of 300 moments that were collected
by systematically varying these parameters, we embarked on 
a large-scale collection
of \nummoments\  happy moments. We also experimented with two platforms: Mechanical Turk and Crowd Flower. We did not notice a significant difference in the quality or content of the moments so we used Mechanical Turk. 

\begin{figure}[H]
\fbox{
\small 
\parbox{0.45\textwidth}{
\underline{\bf Instructions for workers ($I_1$):}\\
\noindent
What made you happy today? Reflect on the {\color{red} past 24 hours}, and recall three {\color{red} actual events} that happened to you that made you happy. Write down your happy moment in a complete sentence. Write three such moments. 

Examples of happy moments we are looking for:
\begin{itemize}
\vspace{-2mm}
\item
I went for a run in the neighborhood. I enjoyed the perfect weather.
\vspace{-2mm}
\item
The offsite with colleagues was great fun. We had stimulating discussions.
\vspace{-2mm}
\item
My son gave me a big hug in the morning when I woke him up.
\vspace{-2mm}
\item
I finally managed to make 40 pushups.
\vspace{-2mm}
\item
I enjoyed watching the sunset on the porch.
\vspace{-2mm}
\end{itemize}

\small
Examples of happy moments we are {\color{red} NOT} looking for (e.g., events in distant past, partial sentence):
\begin{itemize}
\vspace{-2mm}
\item The day I married my spouse
\vspace{-2mm}
\item My Dog
\vspace{-2mm}
\end{itemize}
}}
\normalsize
\caption{Instructions $I_1$ with positive examples}
\label{fig:instructions}
\end{figure}

\subsection{Instructions for workers}
\label{sec:instructions}
Following~\cite{seligman2005positive} who developed a questionnaire that included a question that asked for 3 good things that went well each day, 
we asked our MTurk workers for 3 happy moments that happened to them in the past 24 hours. 
To minimize the bias we introduce through our instructions, we carefully analyze the effect our examples of happy moments
have on the way crowdsource workers report their moments.
In our first batch of crowdsourcing, we gave concise instructions
and positive examples of what we believed are legitimate happy moments (see Figure~\ref{fig:instructions}).

Upon collecting our first batch of 300 happy moments, we tabulated the top 20 most frequent words (nouns/verbs/adjectives) that occur in the  happy moments. It was surprising to 
observe that the words used in the positive examples often appear in the top 20 most frequent words.
(See the first two columns of top of Table~\ref{table:instructions}). 
For example, the bold words ``morning'', ``enjoy'', ``woke'', ``son'', and 
``great'' (excluding ``went''), which are words we use in our positive examples, appear highly among 
the top 20 words.
Consequently, we experimented with an instruction set {\em without}
the positive examples ($I_2$) and collected another 300 happy moments. The top of Table~\ref{table:instructions} shows the top 20 most frequent words.
When we compare the moments collected under $I_1$ and $I_2$, it is evident that the bold words no longer appear highly ranked.

The bottom of Table~\ref{table:instructions} shows a quantitative 
representation of the {\em framing effect} \cite{Tversky:1981cd} of the $I_1$ instruction set.
Here, $\cal V$ denotes the set of all words (no duplicates and no stopwords) obtained from our 5 positive examples 
in the $I_1$ instruction set. 
We counted the number of times words in $\cal V$ occur in the happy moments obtained from $I_1$ instruction and, respectively, $I_2$ instruction. In addition, we also counted the number of times words outside $\cal V$ occur in the respective
batches of happy moments.

\begin{table}[H]
\centering
\small
\begin{tabular}{lc|lc}
\hline
{\bf $I_1$ } &  {\bf Count} & {\bf $I_2$} &  {\bf Count} \\
\hline
\hline
             got &  37 &                 made &  42 \\
            went &  34 &                 long &  23 \\
            made &  27 &                  got &  23 \\
     {\bf great} &  25 &               family &  18 \\
   {\bf morning} &  21 &                 went &  18 \\
         friends &  20 &                 team &  17 \\
            nice &  16 &             favorite &  17 \\
           super &  16 &                   go &  16 \\
             may &  15 &                 work &  15 \\
            bowl &  14 &                 game &  15 \\
   {\bf enjoyed} &  14 &                movie &  14 \\
       {\bf son} &  14 &        {\bf morning} &  14 \\
            work &  14 &               friend &  14 \\
          really &  14 &                first &  13 \\
          family &  14 &               bought &  13 \\
             see &  13 &               dinner &  12 \\
          dinner &  13 &               really &  12 \\
      {\bf woke} &  13 &                  get &  12 \\
        daughter &  12 &              friends &  11 \\
          coffee &  12 &                  saw &  11 
\end{tabular}

\vspace{5mm}
\begin{tabular}{lcc}
\hline
{} & {\bf \# words in batch} & {\bf \# words in batch} \\
{} & {\bf from $\cal V$} & {\bf not from $\cal V$} \\
\hline
\hline
$I_1$ & 213 & 3,115 \\
$I_2$ & 126 & 3,452
\end{tabular}
\caption{The top shows the most used words in happy moments for instruction sets $I_1$ and $I_2$. The bottom shows the frequencies in $I_1$ and $I_2$ w.r.t. $\cal V$, the vocabulary in our positive examples.}
\label{table:instructions}
\end{table}

A $\chi^2$ analysis \cite{IIR} shows that the presence of the 5 positive examples in $I_1$ does affect the word usage of 
workers. 
Specifically, our null hypothesis is that the word usage of the happy moments is independent of whether
the instruction set contained positive examples or not.
The $\chi^2$-test rejected the null hypothesis with $p$-value $< 0.001$. 
Hence, we conclude that MTurk workers were influenced by the positive 
examples in our instructions when reporting their happy moments and 
decided against positive example sentences in the instructions for collecting happy moments.
It is interesting to note that the bottom of Table~\ref{table:instructions} shows also that 
the vocabulary of happy moments from $I_1$ instruction set is significantly minimized,
3,328 total words used in the $I_1$ batch versus 3,578 in the $I_2$ batch.


From this analysis, we concluded that we 
should avoid using positive examples in our instructions.
We also experimented with instructions that do not include negative examples. However,
apart from some reduction in the number of low-quality happy moments, 
we did not detect significant differences between happy moments that are collected from
instructions with or without negative examples. Hence, we included negative examples in our
instructions for the workers.

\section{A Spell-Corrector for \hdb}
\label{sec:spell}
Here, we discuss the details of the spell-correction algorithm that we have created for \hdb.
Our main goal is to fix as many typos while introducing as little error as possible. To this end,
we have decided to focus on a small set of corrections: typos that are within a Levenshtein distance
of 1 of a valid word (i.e., one deletion, insertion, transposition, or replacement of a letter or a space).

The spell-correcting algorithm starts by finding the set of words
within edit distance 1 of a typo and computes a confidence score $C(w)$,
for each word $w$ which we defined as
$C(w) = \textnormal{log}(f_w)$
where $f_w$ is the frequency of the word $w$. If $w$ consists of two words (which occurs with replacement or
insertion of a space character) then $f_w$ is the lower frequency of the two words. We calculate these frequencies
using a corpora which consists of Norvig's corpus and the portion of \hdb\ that has no spelling errors. We observed
that resulting corrector was biased toward splitting words into two, for example ``outtdoors'' was being replaced with
``out doors'' instead of ``outdoors''. This is because shorter words occur more frequently. For the same reason,
words were often being replaced by an incorrect but shorter alternative, for example ``helpd'' being replaced with
``help'' (instead of ``helped''). In an attempt to solve this problem, we refined the confidence score by adding two
additional parameters: $0\leq s \leq 1$, to discount the confidence of replacements with an inserted space, and $l$,
to increase the confidence of longer words. This updated confidence score $C(w)$ can be written as
\[
C(w) = \textnormal{log}(f_w)s^{b(w)} + l \times \textnormal{len}(w)
\]
\noindent
where $b(w)$ is a simple indicator function which returns $0$ if $w$ consists of two words and returns $0$ otherwise.
Note that we are using the logarithmic frequency in our definition. The hypothesis is that shorter words occur
exponentially more often than long words on average, in which case computing confidence as a function of the logarithmic
frequency would yield better results. We also observed this effect in practice. The last step is to tune the parameters
$s$ and $l$. To tune these parameters, we took a random sample of 100 spelling errors from our data and manually
made corrections. We then perform a grid search over possible values for $l$ and $s$. Our experiment suggests that
conservative values for $l$ and $s$ are $0.5$ and $0.05$, respectively. With these values, in a random sample of $3,000$
happy moments, all 74 detected typos were either corrected appropriately or left as is.

We applied our spell-corrector on the entire corpus, which automatically replaced $1,568$ words. The remaining typos were
corrected by workers on CrowdFlower, as well as internal workers in our lab. Each word was evaluated by two judges, and if
they agreed, the result was automatically applied. There were less than 500 words where the judges did not agree. In this case,
we defaulted to the worker with higher confidence rating (our internal lab workers). In the remaining cases where the confidence
ratings were equal, we left the word alone if either judge was unsure, otherwise chose a suggestion at random (in the majority
of these cases, the answers varied in usage of space, punctuation, or capitalization, and not in content). In total, $2,218$
happy moments were modified with this method.

\end{document}